\documentclass[journal]{IEEEtran}

\usepackage{cite}

\ifCLASSINFOpdf
  \usepackage[pdftex]{graphicx}
  \graphicspath{{./pics/}}

\else

\fi

\usepackage[
singlelinecheck=false 
]{caption}

\ifCLASSOPTIONcompsoc
  \usepackage[caption=false,font=normalsize,labelfont=sf,textfont=sf]{subfig}
\else
  \usepackage[caption=false,font=footnotesize]{subfig}
\fi
\usepackage{subfloat}
\usepackage{caption}

\usepackage{amsmath}
\usepackage{amssymb}
\usepackage{amstext}
\usepackage{amsfonts}
\usepackage{url}
\usepackage{bbm}
\usepackage{color}

\usepackage{tablefootnote}

\usepackage{float}

\usepackage{color}
\definecolor{red}{RGB}{255,0,0}

\begin{document}
%
\title{Integrating Scene Text and Visual Appearance for Fine-Grained Image Classification}

\author{Xiang~Bai,
        Mingkun~Yang,
        Pengyuan~Lyu,
        Yongchao~Xu,
        and Jiebo~Luo
    \thanks{Xiang Bai, Mingkun Yang, Penyuan Lyu, and Yongchao Xu are with the School of Electronic Information and Communications, Huazhong University of Science and Technology (HUST), Wuhan, 430074, China. Email: \{xbai, yangmingkun, lvpyuan, yongchaoxu\}@hust.edu.cn.}
    \thanks{Yongchao Xu is also with EPITA Research and Development Laboratory, 14-16 rue Voltaire, FR-94270 Le Kremlin-Bicetre, France}
    \thanks{Jiebo Luo is with the Department of Computer Science, University of Rochester, Rochester, NY 14627 USA. Email: jiebo.luo@gmail.com.}
}

\markboth{??? of \LaTeX\ Class Files,~Vol.~?, No.~?, ?~?}%
{Shell \MakeLowercase{\textit{et al.}}: Bare Demo of IEEEtran.cls for IEEE Journals}


\maketitle

\begin{abstract}
Text in natural images contains rich semantics that are often highly relevant to objects or scene. In this paper, we focus on the problem of fully exploiting scene text for visual understanding. The main idea is combining word representations and deep visual features into a globally trainable deep convolutional neural network. First, the recognized words are obtained by a scene text reading system. Then, we combine the word embedding of the recognized words and the deep visual features into a single representation, which is optimized by a convolutional neural network for fine-grained image classification. In our framework, the attention mechanism is adopted to reveal the relevance between each recognized word and the given image, which further enhances the recognition performance. We have performed experiments on two datasets: Con-Text dataset and Drink Bottle dataset, that are proposed for fine-grained classification of business places and drink bottles, respectively. The experimental results consistently demonstrate that the proposed method  combining textual and visual cues significantly outperforms classification with only visual representations. Moreover, we have shown that the learned representation improves the retrieval performance on the drink bottle images by a large margin, making it potentially useful in product search.        
 
\end{abstract}


\IEEEpeerreviewmaketitle

\section{Introduction}

Combining multi-modal features for the recognition or index of visual data is the inevitable way to automatically understanding image/video content in the recent years. The multi-modal visual data are often correlated and complementary to each other in visual understanding, especially when they originate from the same source. In this paper, we are concerned on how to efficiently integrate visual and textual cues for fine-grained image classification with convolutional neural networks. The goal of fine-grained classification is to assign the labels to the images having subtle differences in visual appearance such as animal species, product types, and place types. Such a problem is quite challenging, as relevant differences that are not obvious may not be accurately distinguished by a typical classification model. Even for human perception, domain specific knowledge is often required for fine-grained classification. 

Scene text, which frequently appears in natural scenes including road signs, street views, product packages, and license plates, often possesses rich and precise meanings that are highly related to semantics of the object or scene in the same image. For instance, texts on the buildings or wrapping bags are quite useful for distinguishing among their categories. With the recent developments in text detection and recognition in the wild \cite{YeAndDoermann,ZhuYaoBai}, it has been shown that textual cues are very beneficial to fine-grained classification, especially in the classification of business places such as bakery, cafe and bookstore \cite{karaoglu2016words}. 

\begin{figure}[h]
  \centering
  \includegraphics[width=\linewidth]{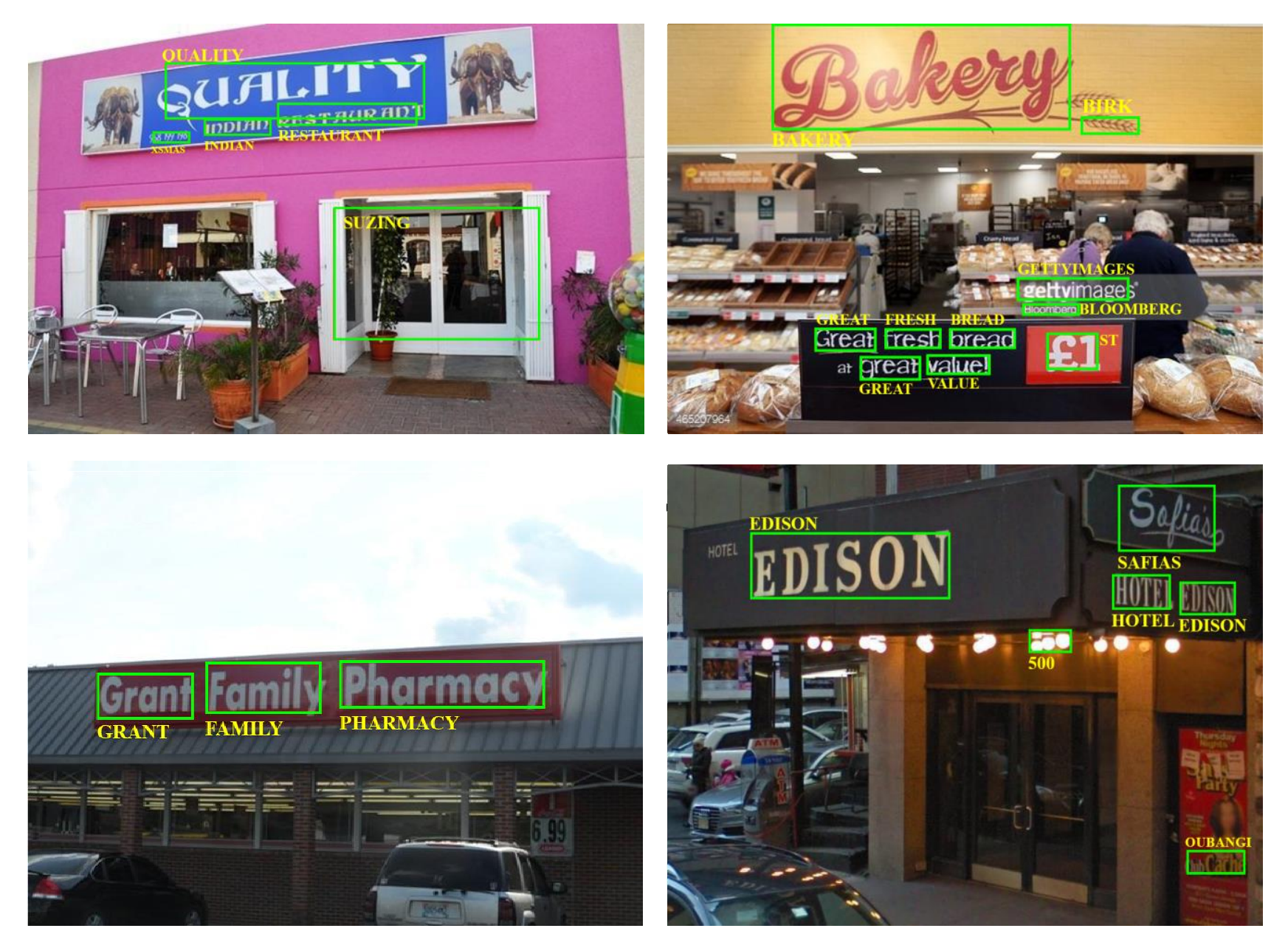}
  \caption{The recognized words in some scene images obtained by a text reading system.}
   \label{fig:RecWord}
\end{figure}

Scene text detection and recognition has been an active research area in both computer vision and document analysis communities. A large number of novel approaches are proposed for the localization and recognition of the text embodied in natural images and videos. Such approaches enable us to automatically extract the text information, which can be considered as an additional information for image classification. In this paper, the recognized words in natural scenes are used as a kind of input features for an image classification model. However, given an image, not every recognized word must have the relation to it. As shown in Figure~\ref{fig:RecWord}, the recognized words such as `restaurant', `bakery', `pharmacy', `bread', `hotel' are very relevant to the scene semantics, but some words like `grand', `great', `edison' don't have the direct connection to them. This requires a model to measure the relevance of each recognized word to the semantics of objects or scenes, in order to make full use of textual information in image classification. 
It should be mentioned that the existing robust reading system can not guarantee that all the words embodied in the images can be correctly detected and/or recognized. Yet, even under such a situation, the recognized words are still very helpful for image classification as demonstrated by our experiments.  

To the best of our knowledge, there has been little research on combining textual and visual cues for fine-grained image classification. Therefore, it is imperative to develop such technologies for many fine-grained classification tasks such as the recognition or indexing of films, products, logos, places, athletes, advertisements, and so on, to fulfill the needs of many real-world applications. Since such scenarios often contain texts that are highly relevant and complementary to the visual appearances of objects or scenes, the text semantics are quite helpful for enhancing the performances of fine-grained classification.

\begin{figure*}[h]
    \centering
    \includegraphics[width=\linewidth]{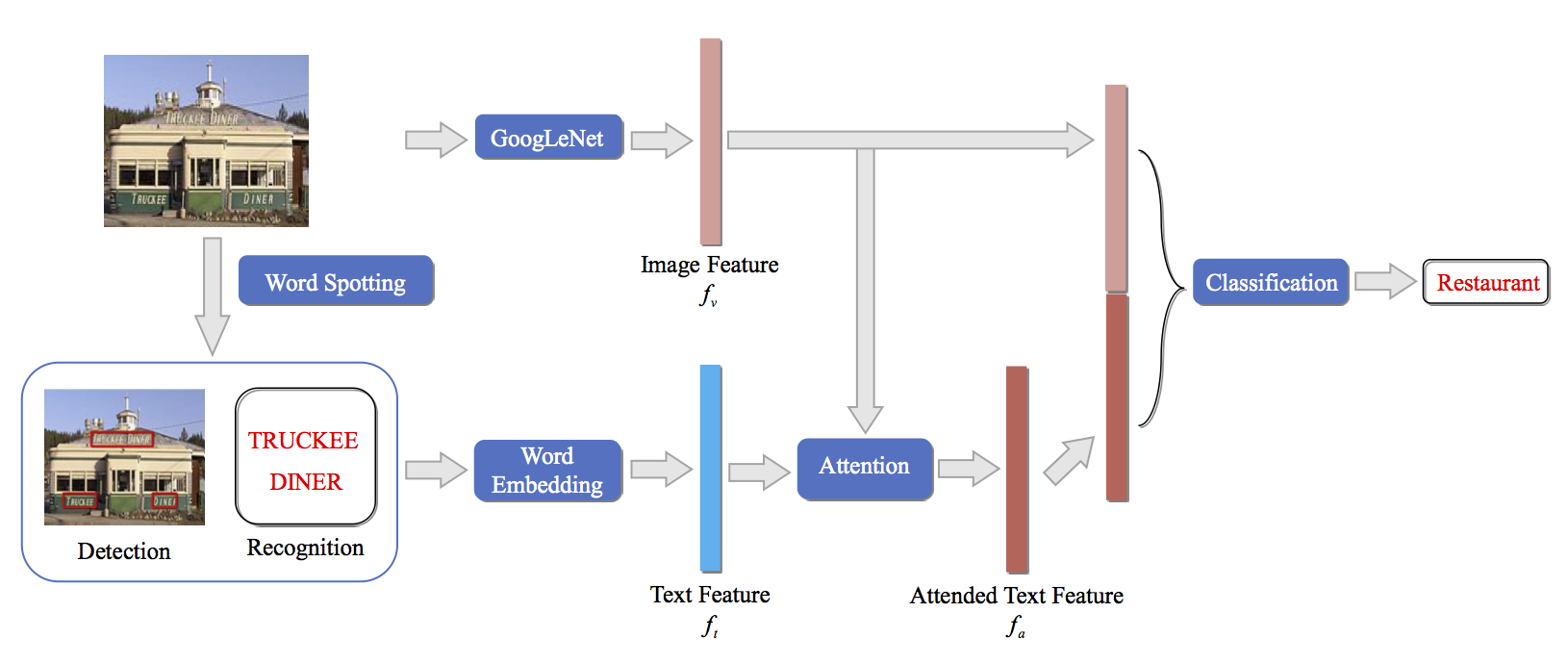}%
    \caption{Pipeline of the proposed model. Given an input image $I$, the GooLeNet \cite{szegedy2015going} CNN extracts a $1024$-dimensional feature vector as the visual representation. A state-of-the-art text detector \cite{liao2016textboxes} is used to localize words in the image, followed by a text recognizer \cite{shi2016end}. The recognized words are transformed into the textual representation using word embedding. Then the method uses semantic attention mechanism to filter out some noisy words. At last, the multi-modal features are fused for fine-grained classification.}
    \label{fig:overall-arch}
\end{figure*}

Recently, deep neural networks have been the mainstream techniques for visual understanding, thanks to their great successes in various vision tasks. In this paper, we focus on designing a proper deep neural network architecture for fine-grained classification, which is able to efficiently fuse the semantics of text and visual cues. We firstly perform word detection and recognition in the images with a recent state-of-the-art scene text reading system \cite{liao2016textboxes}. Then, word embedding is adopted to transfer the recognized words into a vector space. Finally, the word features are used as an extra input to a typical convolutional neural network \cite{szegedy2015going} for fine-grained classification. The proposed unified framework for integrating text and visual features is illustrated in Figure \ref{fig:overall-arch}, which is end-to-end trainable.

Our contributions in this paper are three  folds:

1) We present a simple yet effective pipeline, using the state-of-the-art components for scene text and visual recognition,  for fine-grained image 
classification through combining textual and visual
cues. Our results show that the meanings of recognized words
are often complementary to the visual information, resulting in
the significant improvements on fine-grained classification.

2) We develop an end-to-end trainable neural network
with the attention mechanism for implicitly exploiting the semantic
relationships between textual and visual cues. Such a unified
framework is more elegant and efficient than the existing
alternatives designed for cross-domain representations.

3) We collect and will release a new image dataset consisting of various types of drink bottle images. Our proposed method works well on drink bottle image classification and
retrieval, demonstrating its potential in product search. 

The remainder of the paper is organized as follows. Section~\ref{sec: related works} briefly reviews some related works. In Section~\ref{section: multi-modal features fusion}, we describe in detail the proposed framework on integrating textual and visual cues for fine-grained classification. Section~\ref{sec: experimental results} provides some experimental results, comparisons, and analyses on two datasets. Finally, conclusion remarks and future works are given in Section~\ref{sec: conclusion}.

 

\section{Related Works} \label{sec: related works}
\subsection{Fine-grained Classification}
Recently, fine-grained image classification is an active topic in some domains, such as animal species~\cite{zhang2014part}, plant species~\cite{angelova2013efficient} and man-made objects \cite{krause20133d}. Different from the base-class image classification, fine-grained image classification needs to distinguish images with more subtle differences among object classes. Discerning the subtle differences among similar classes is more challenging. Many recent works rely on relevant parts of fine-grained classes for domain specific knowledge. Examples are the recent works in \cite{yang2012unsupervised,deng2013fine} that use salient parts in images. Semantic object parts are used in \cite{zhang2014part} to isolate the subtle differences. Xiao \textit{et al.} \cite{xiao2015application} used three attention models to first propose candidate patches, then to localize discriminative parts by selecting relevant patches of a specific object. In \cite{zhang2016weakly}, the authors proposed to select many useful parts from multi-scale part proposals of objects in each image without learning expensive object/part detectors. A weakly supervised strategy that using Multi-Instance Learning (MIL) to learn discriminative patterns for image representation is proposed in~\cite{tang2016learning}. Zhang \textit{et al.} \cite{zhang2012pose} develop a pose-normalized representation for image patches to provide a high degree of pose invariance. 

Besides these part-based methods, many other methods have been developed. For example, in \cite{wang2014learning}, the authors adopted a deep ranking model to learn similarity metric directly from images. A multi-stage distance metric learning framework is proposed in \cite{qian2015fine} to learn a more discriminative similarity. Xie \textit{et al.} \cite{xie2015hyper} proposed a novel data augmentation approach to address the scarcity of data and a novel regularization technique to improve the generalization performance. In~\cite{deng2013fine} and~\cite{wilber2015learning}, the authors proposed to include humans in loop to select discriminative features. 

The most related work with our proposal in this paper is the one in~\cite{karaoglu2013text, karaoglu2016words}, where scene texts in the business place images are considered as the domain knowledge. The authors proposed to combine visual and textual cues for fine-grained business place classification. More precisely, they train two one-versus-rest SVM classifiers on visual and respectively text features. Then these two trained SVM kernels are added together to form the final SVM kernel matrix. In this way, the visual and text cues are combined together. In~\cite{karaoglu2013text}, Bag-of-Words are used to represent visual cues. Deep visual cues given by GoogleNet~\cite{szegedy2015going} are used in~\cite{karaoglu2016words}. Sharing the same spirit as part-based methods, the authors proposed in~\cite{karaoglu2016words} to extract 100 EdgeBoxes~\cite{zitnick2014edge}. Then the final visual feature is given by the combination of the GoogleNet visual features for the 100 boxed regions and the GoogleNet visual features for the whole image. In our work, we also propose to integrate both visual and text cues for fine-grained classification. We rely on an attention-based model to select the relevant words and get rid of those irrelevant ones. Hopefully, the combination of the attended text cues and visual feature of the whole image are more discriminative. Besides, as shown in Figure~\ref{fig:overall-arch}, our proposed model is end-to-end trainable, which is more elegant and efficient. 

\subsection{Scene text detection and recognition}
Scene text detection and recognition have been extensively studied in computer vision and document analysis communities \cite{chen2004detecting, epshtein2010detecting, wang2010word, pan2011hybrid, zhao2010text,yi2011text,yin2014robust,yao2012detecting,neumann2010method,wang2011end,yao2014unified,neumann2013scene,wang2012end,bissacco2013photoocr,jaderberg2014deep,jaderberg2016reading,gupta2016synthetic,huang2014robust,huang2013text,lu2015scene,neumann2012real,tian2015text,zhong2016deeptext,he2016aggregating,gomez2016textproposals,calarasanu2016good} (see \cite{YeAndDoermann, ZhuYaoBai} for comprehensive surveys). Since the proposed method mainly leverages the recognized words to improve fine-grained classification, this section briefly revisits the most relevant works about end-to-end word recognition. In the detection stage, connected components \cite{neumann2010method, neumann2013scene, yao2014unified} or sliding windows \cite{wang2011end, wang2012end} are usually used to extract character candidates. Then, the character candidate regions are further verified and grouped into word candidates based on some geometric rules or learning models. In the recognition stage, character-based methods \cite{wang2011end, yao2014unified, mishra2012scene, mishra2012top, bai2016strokelets} and word-based methods \cite{jaderberg2016reading, almazan2014word, shi2016end} are the two mainstream recognition approaches. Recently, following the popular general object detection approaches such as RCNN \cite{girshick2014rich}, Faster RCNN \cite{ren2015faster}, YOLO \cite{redmon2016you}, and SSD \cite{liu2016ssd}, 
the scheme of end-to-end word proposal generation followed by proposal classification has been a trend for word recognition in natural images~\cite{jaderberg2016reading, gupta2016synthetic, zhong2016deeptext, gomez2016textproposals, liao2016textboxes}. In this paper, we adopt a recent algorithm for end-to-end word recognition named TextBoxes \cite{liao2016textboxes}, which is the fastest among the existing text reading systems.

\subsection{Multi-modal Fusion}
In many tasks, information about the same phenomenon can be easily obtained from various sources, such as video, audio, images, and text. 
Generally, information from different modalities is complementary and offers useful knowledge to each other. Multi-modal approaches have shown superior performances over single-modal ones for many tasks~\cite{Bhatt2011Multimedia}.
For example, in \cite{karaoglu2016words, karaoglu2013text}, the authors proposed to combine visual and textual cues to improve fine-grained classification of business place images. Similarly, Lu \textit{et al.} \cite{lu2014text} discussed some methods that combine visual cues and video captions to improve video classification. In~\cite{rusinol2014multimodal}, the authors presented a page classification application in a banking workflow by merging visual and textual descriptions.

The main challenge of multi-modal approaches lies on how to optimally combine the multi-modal information. The combination can take place at different levels. The most widely used strategy is to fuse the information at the feature level, which is known as early fusion. This kind of system merges the extracted features using integration mechanisms, such as concatenation \cite{kiela2014learning}, neural network architectures \cite{andrew2013deep}, etc. Another combination strategy is at decision level or late fusion \cite{snoek2005early}, which merges multiple modalities in the semantic space. A hybrid combination of the early and late fusion is also proposed \cite{d2007hybrid}.

Following \cite{karaoglu2016words, karaoglu2013text}, we also propose to combine visual and text cues to improve fine-grained classification. We combine the visual cues and textual cues at the feature level.

\subsection{Attention Mechanism}
Attention mechanism is widely used in natural language processing, such as speech recognition \cite{bahdanau2016end}, machine translation \cite{bahdanau2014neural}, and dialog management \cite{hori2016dialog} to improve the network's ability of finding relevant features from corresponding input parts. Recently, visual attention attracts much attention in computer vision society. Ba \textit{et al.} \cite{ba2014multiple} presented an attention-based model for recognizing multiple objects in images. In~\cite{gregor2015draw}, the authors proposed a novel network which combines the attention mechanism and an auto-encoding framework for iterative image construction. The attention mechanism is applied to video description in~\cite{hori2017attention}, and \cite{xu2015show, you2016image} studied the spatial attention for image captioning. 

In our case, even though a powerful text reading system TextBoxes~\cite{liao2016textboxes} is used to spot words in natural images, there are still some errors (see Figure~\ref{fig:RecWord} due to large variations in both foreground texts and background objects, and uncontrollable lighting conditions). To reduce the influence of false detections and irrelevant words, we introduce attention mechanism to automatically analyze the correlation between texts and images. Hopefully, the attention mechanism helps to select most related words in images for the fine-grained classification.


\section{Fine-grained classification integrating text and visual appearance} \label{section: multi-modal features fusion}

\subsection{Proposed pipeline}
In this paper, we present an end-to-end neural-network-based model for improving the fine-grained classification by combining visual and text features in the image. The proposed network structure is depicted in Figure~\ref{fig:overall-arch}. It consists mainly of three parts: visual and text feature extraction, semantic attention mechanism for noisy text filtering, and multi-modal feature fusion. The first part aims at transforming each input source into a feature representation. The second one is dedicated to reduce noisy texts brought in during text spotting. The last part combines different modality representations into a single representation. The whole framework is end-to-end trainable. In the following, we detail the main components of the proposed method.

\subsection{Visual Representation} \label{sssec:visual representation}
In computer vision tasks, convolutional neural networks have become a popular method to extract image features. In~\cite{sharif2014cnn}, it has been shown that CNN models trained on a large image dataset can learn common features, which can be shared in many different tasks. In~\cite{karaoglu2016words}, the authors have also shown that fine-tunning from a trained model can obtain better performance than training from scratch in some tasks. Based on this property, we use GoogleNet \cite{szegedy2015going} trained on millions of ImageNet \cite{russakovsky2015imagenet} images as initial CNN model and fine tune its parameters with other parts of our model in an end-to-end way. Given an image $I$, we use the output of the last average pooling layer of the fine tuned GoogLeNet as the representation of visual features $f_v$, a 1024-dimensional feature vector.


\subsection{Text Representation} \label{sssec:text representation}
Word2Vec \cite{mikolov2013efficient} and GloVe \cite{pennington2014glove} are two popular word embedding algorithms to construct vector representations for words. These vector-based models represent words in a continuous vector space where semantically similar words are embedded nearby each other. Pennington \textit{et al.} \cite{pennington2014glove} noted that GloVe consistently outperforms word2vec for the same corpus, vocabulary, window size and training time. So we adopt GloVe in this paper to transform recognized texts into text features. More specifically, we represent each spotted word $T_i$ by the pre-trained 300-dimensional word vector $f_{t_i}$ of GloVe.
For an image with $N$ spotted words, this results in a text feature $f_t$ whose size is $300 \times N$. To train the network efficiently with batch, we fix the size $N$ with the maximum number of spotted words in each training image denoted as $N_{\max}$. For the training and testing images having less than $N_{\max}$ spotted words, we use zero padding. For the testing images having more than $N_{\max}$ spotted words, we sort the words in decreasing order of their recognition scores and use the first $N_{\max}$ ones to form the text representation. Finally, the text information in each image is represented by a 
$300 \times N_{\max}$ feature.

\subsection{Attention Unit}\label{subsec:attention}
\begin{figure}[t]
  \centering
  \includegraphics[width=\linewidth]{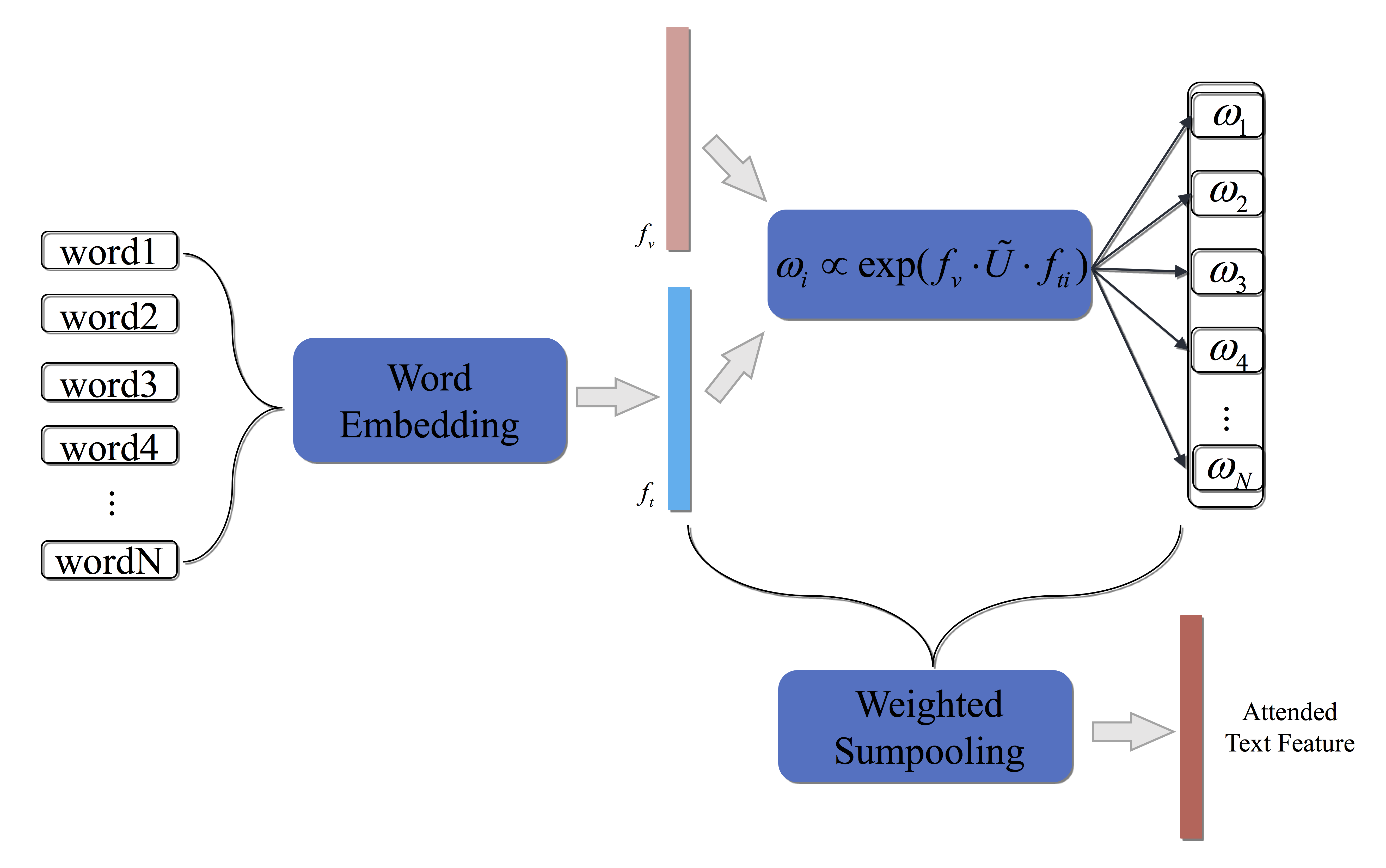}
  \caption{The architecture of the proposed semantic attention model. Visual features $f_v$ and text representation $f_t$ are injected into a bilinear function for attention. Different words result in different attention weights, they are fused together by weighted sum-pooling for more discriminative representation.}
  \label{fig:attention model}
\end{figure}

Generally, not all the words in an image must have semantic relations to the scene. Some words may have strong correlations with the image, others may be not relevant at all. Consequently, different words should have different weights representing their correlations with the image. For example, it is reasonable that a word being more related to the image or having a high recognition score should have a large weight, and vice versa. Besides, the text reading system could make some mistakes and result in some noisy texts, which should be discarded. These noisy texts should have a very small weight. Inspired by the work in \cite{you2016image} which proposed a semantic attention mechanism to selectively attend on rich semantic attributes detected from the image, we apply a similar semantic attention mechanism to assign different weights over words in our case. The architecture of the proposed attention model is depicted in Figure~\ref{fig:attention model}.

In the attention unit, a bilinear function is used to estimate the weight $\omega_i$ for each spotted word $T_i$ in the $N_{\max}$ words reflecting the image's text feature. This weight $\omega_i$ measures the correlation between the spotted word $T_i$ and the image. It is given by:
\begin{equation}
\omega_i\propto exp(f_v^T \cdot \tilde{U} \cdot f_{t_i}),
\end{equation}
where the exponent is taken to normalize over all words $\mathbb{T} = \{T_i\}$ in a softmax fashion, and $\tilde{U}$ is the $1024$ $\times$ $300$ bilinear parameter matrix 
We then use weighted sum-pooling to fuse all the $N_{\max}$ word features. This fused word feature named attended text representation $f_a$ is given by:
\begin{equation}
f_a = \sum_{i=1}^{N_{\max}} w_i\cdot f_{t_i}.
\end{equation}


\subsection{Multi-Modal feature fusion}

Multi-modal feature fusion has shown superior performance over unimodal approaches in many tasks~\cite{deng2014deep}. A common multi-modal feature fusion method takes two or more modalities to form a preferable feature. In our problem, visual and attended text features provide important and complementary information about the image. The fine-grained classification would benefit from a combination of these two features.

Since the statistical properties are usually not shared across modalities, a transformation of features is required before their fusion. We use a 512-dimensional fully connected layer to map the visual feature into a suitable space $f_{v'}$ having an approximate dimensionality as the attended text feature. Moreover, as shown in many applications, batch-normalization \cite{ioffe2015batch} could improve the performance. In this paper, we also adopt this strategy to normalize visual and text features before the multi-modal fusion. Since the main focus of this paper does not lie on how to combine these two features, we simply concatenate them as the final multi-modal feature $f_c = [f_{v'}, f_a]$.
\subsection{Supervision Over Different Features} \label{sssec:supervison over different features}

Multitask Learning (MTL)~\cite{caruana1998multitask} is an inductive transfer mechanism which leverages the domain-specific information to improve the performance of generalization.
It relies on training multi-tasks in parallel using a shared representation, which enables them help each other to improve their performances.
Inspired by MTL, we add two extra classifiers on visual feature $f_v$ and respectively on attended text feature $f_a$, in parallel with the classifier based on multi-modal feature $f_c$. Simply put, we train in parallel three classifiers for fine-grained classification: 1) a classifier based on combined visual and attended text feature $f_c$; 2) a classifier using only visual feature $f_v$; 3) a classifier relying on attended text feature $f_a$. Note that for the last classifier trained on $f_a$ only, we add an extra class representing images without recognized text to the set of ground truth categories. All the three classifiers are built in a softmax fashion. The two extra supervisions on $f_v$ and respectively $f_a$ make them more discriminative when used alone, so is their combination. Consequently, this would improve the performance of the proposed classifier integrating both visual and text appearance.



For an image $I$ having ground truth label $Z_i$, let $\textbf{W}$ and ${\bf {\Phi}}_i$ be the parameters of the shared network and respectively the parameters related to the $i$-th classifier ($i \in \{1, 2, 3\}$). We train the $i$-th classifier by minimizing the loss function ${\cal L}_i$(\textbf{W},\textbf{$\Phi_i$}) given by:
\begin{equation}
{\cal L}_i(\textbf{W},{\bf \Phi_i}) = -\sum_{k=1}^{K_i} \textbf{1}(Z_i=k) \log Pr(k \mid I,\mathbb{T}; \textbf{W},{\bf \Phi_i}),
\end{equation}
where $K_i$ is the number of categories for the $i$-th classifier, and $Pr(k \mid I,\mathbb{T}; \textbf{W},\textbf{$\Phi_i$}) \in$ [0, 1] is the predicted score given by the classifier measuring the probability of image $I$ belonging to category $k$. Note that $K_1$ and $K_2$ are the number of categories in training set. Since the third classifier is trained on attended text feature with an extra class representing images without spotted text, $K_3 = K_1 + 1$.

Let {\bf $\O$} = $\{{\bf {\Phi}}_i \mid i = 1,2,3\}$ denotes the parameters of the three classifiers, then the final loss function is simply given by:
\begin{equation} \label{fuse_loss}
{\cal L}(\textbf{W}, {\bf \O}) = \sum_{i=1}^{3} {\beta}_i {\cal L}_i(\textbf{W},{\bf \Phi_i}),
\end{equation}
where ${\beta}_i$ is the loss weight for the $i$-th classifier. We minimize this loss function ${\cal L}(\textbf{W}, {\bf \O})$ via standard (back-propagation) stochastic gradient descent (SGD):
\begin{equation}
(\textbf{W},\textbf{$\O$})* = \arg\min({\cal L}(\textbf{W}, \textbf{$\O$}))
\end{equation}
%

\section{Experimental results} \label{sec: experimental results}
\subsection{Experimental setup}

\subsubsection{\textbf{Datasets}} \label{sssec:datasets}
We evaluate the proposed method on two datasets: \textit{Con-Text dataset} proposed in \cite{karaoglu2013text} and our self-collected \textit{Drink Bottle dataset}. The Con-Text dataset is dedicated for fine-grained classification of business places, e.g., Cafe, Bookstore, and Pharmacy. It is composed of 24,255 images classified into 28 categories. All the images in this dataset are split into three folds, where two folds are used for training and the other one is used for testing. Each type of experiment on this dataset are repeated three times using respectively the three possible training and testing settings. The average performance is used as the reported results. 

\begin{figure*}[h]
    \centering
    \captionsetup[subfloat]{justification=centering}
    \subfloat[Root Beer]{\includegraphics[width=1.4in, height=1.3in]{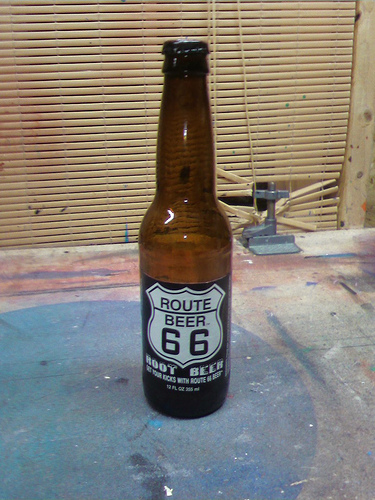}%
        \label{fig_first_case}}
    \hfill
    \subfloat[\scriptsize{Ginger ale}]{\includegraphics[width=1.4in, height=1.3in]{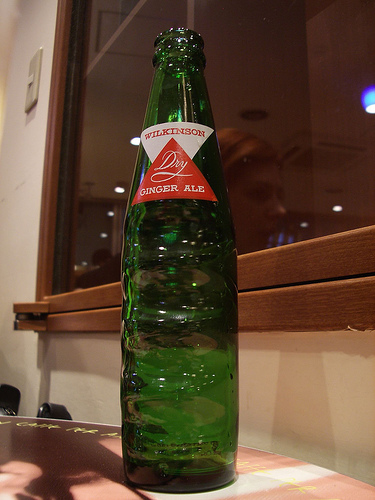}%
        \label{fig_second_case}}
    \hfill
    \subfloat[\scriptsize{Coca Cola}]{\includegraphics[width=1.4in, height=1.3in]{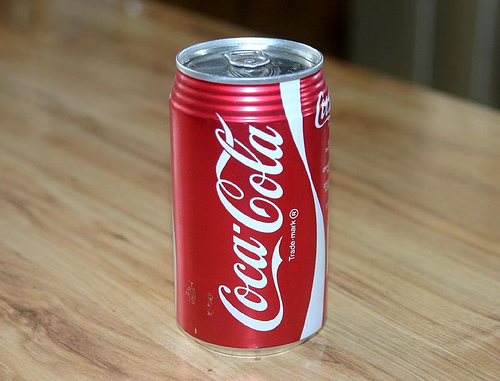}%
        \label{fig_third_case}}
    \hfill
    \subfloat[\scriptsize{Pepsi Cola}]{\includegraphics[width=1.4in, height=1.3in]{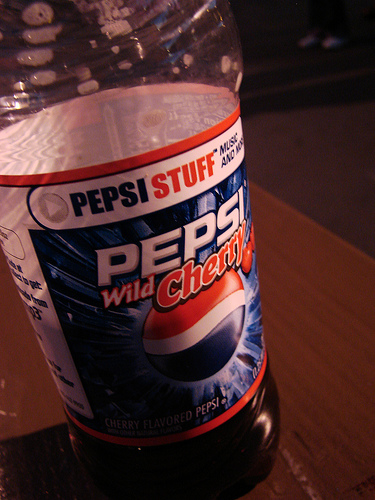}%
        \label{fig_fourth_case}}
    \hfill
    \subfloat[\scriptsize{Cream Soda}]{\includegraphics[width=1.4in, height=1.3in]{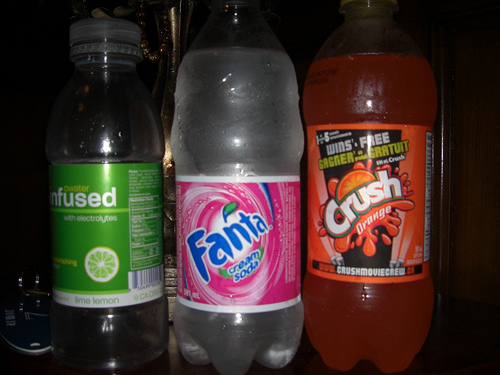}%
    	\label{fig_five_case}}
    \hfill
    \subfloat[\scriptsize{Egg Cream}]{\includegraphics[width=1.4in, height=1.3in]{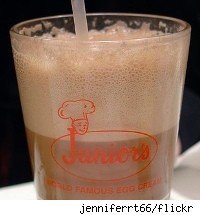}%
    	\label{fig_six_case}}
    \hfill
    \subfloat[\scriptsize{Birch Beer}]{\includegraphics[width=1.4in, height=1.3in]{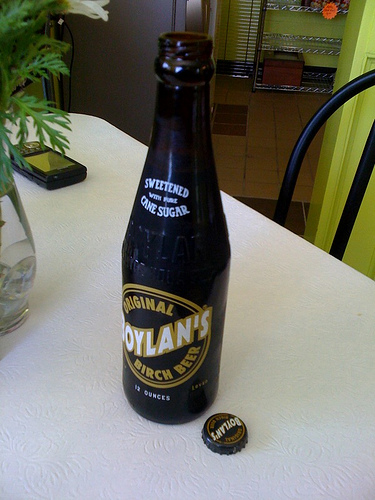}%
    	\label{fig_seven_case}}
    \hfill
    \subfloat[\scriptsize{Tonic}]{\includegraphics[width=1.4in, height=1.3in]{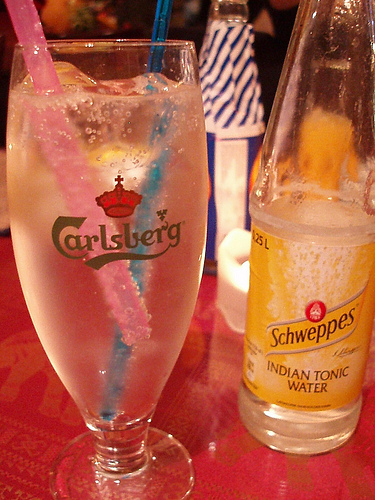}%
    	\label{fig_eight_case}}
    \hfill
    \subfloat[\scriptsize{Sarsaparilla}]{\includegraphics[width=1.4in, height=1.3in]{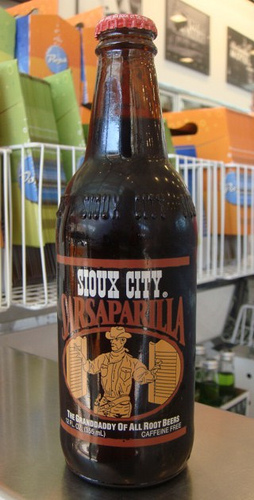}%
        \label{fig_nine_case}}
    \hfill
    \subfloat[\scriptsize{Orange Soda}]{\includegraphics[width=1.4in, height=1.3in]{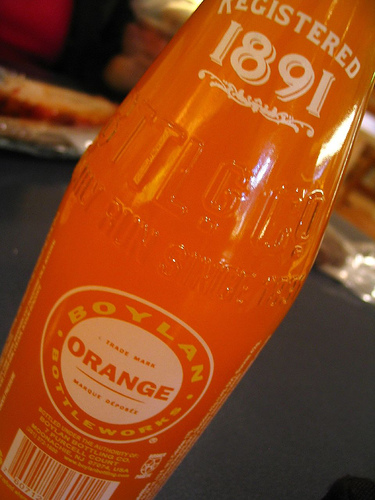}%
        \label{fig_ten_case}}
    \hfill
    \subfloat[\scriptsize{Pulque}]{\includegraphics[width=1.4in, height=1.3in]{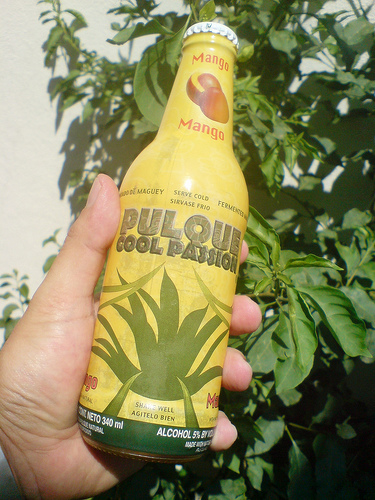}%
        \label{fig_eleven_case}}
    \hfill
    \subfloat[\scriptsize{Kvass}]{\includegraphics[width=1.4in, height=1.3in]{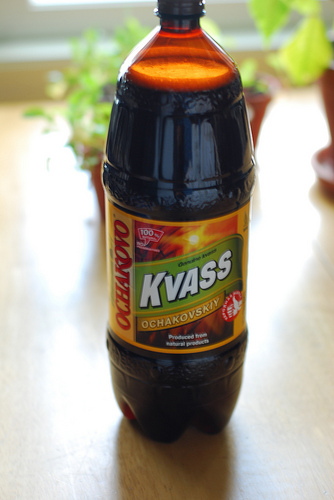}%
        \label{fig_twelve_case}}
    \hfill
    \subfloat[\scriptsize{Bitter}]{\includegraphics[width=1.4in, height=1.3in]{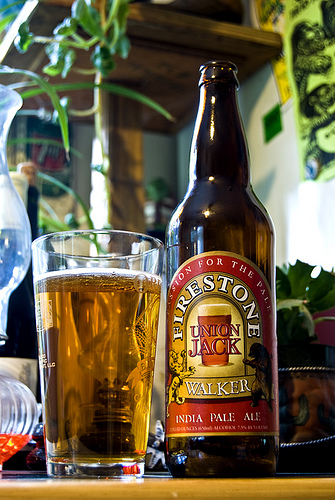}%
    	\label{fig_thirteen_case}}
    \hfill
    \subfloat[\scriptsize{Guinness}]{\includegraphics[width=1.4in, height=1.3in]{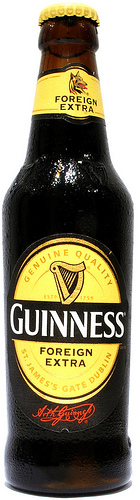}%
    	\label{fig_fourteen_case}}
    \hfill
    \subfloat[\scriptsize{Ouzo}]{\includegraphics[width=1.4in, height=1.3in]{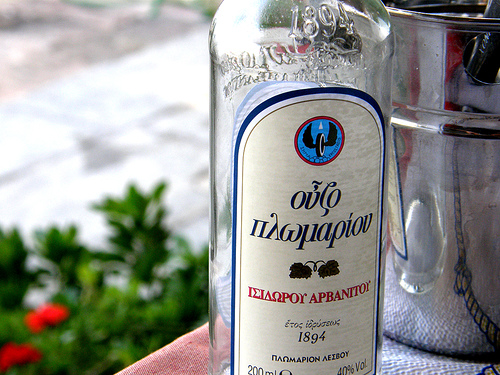}%
    	\label{fig_fifteen_case}}
    \hfill
    \subfloat[\scriptsize{Slivovitz}]{\includegraphics[width=1.4in, height=1.3in]{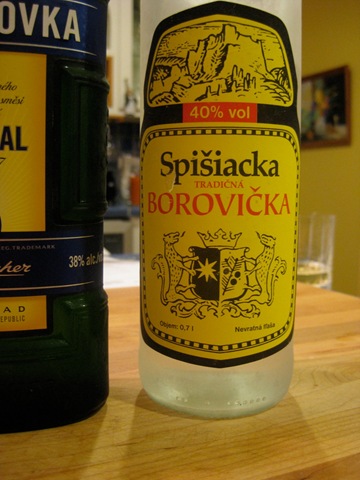}%
    	\label{fig_sixteen_case}}
    \hfill
    \subfloat[\scriptsize{Drambuie}]{\includegraphics[width=1.4in, height=1.3in]{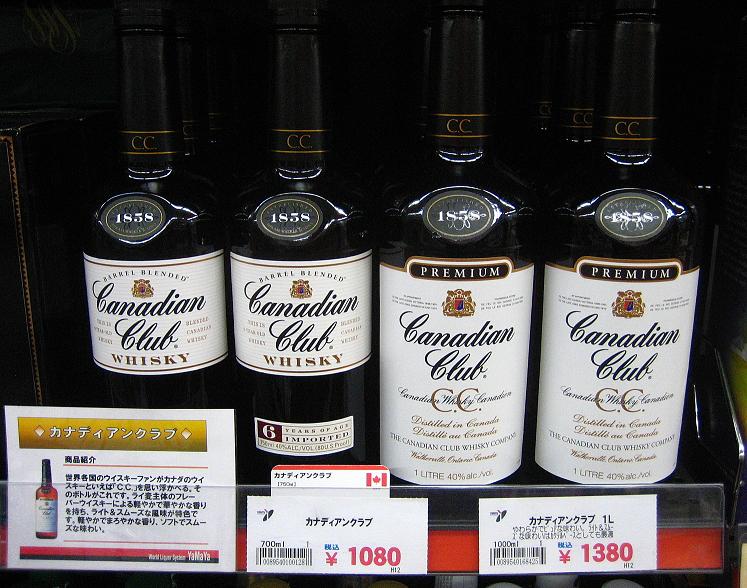}%
    	\label{fig_seventeen_case}}
    \hfill
    \subfloat[\scriptsize{Vodka}]{\includegraphics[width=1.4in, height=1.3in]{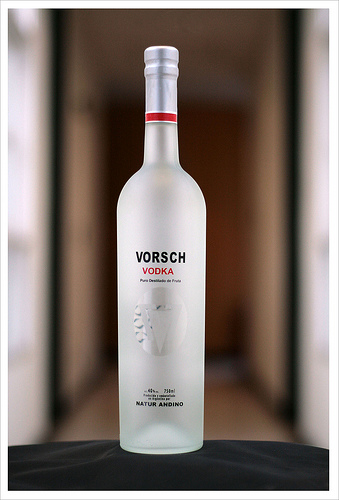}%
    	\label{fig_eighteen_case}}
    \hfill
    \subfloat[\scriptsize{Chablis}]{\includegraphics[width=1.4in, height=1.3in]{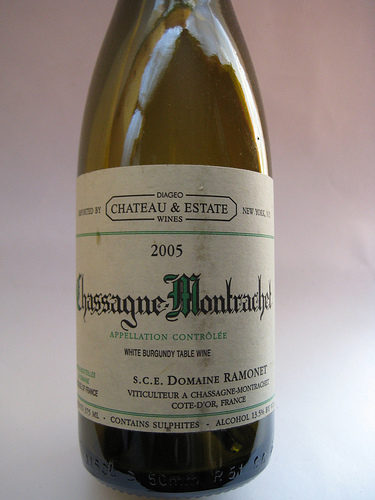}%
    	\label{fig_nineteen_case}}
    \hfill
    \subfloat[\scriptsize{Sauterne}]{\includegraphics[width=1.4in, height=1.3in]{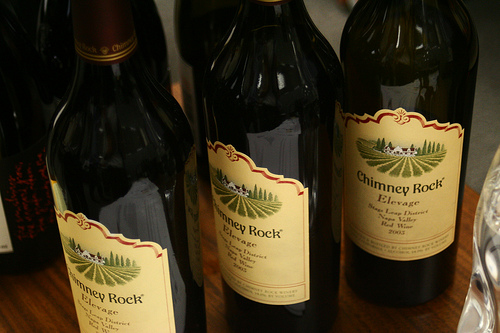}%
    	\label{fig_twenty_case}}
    \caption{Some image samples in the Drink Bottle dataset. They have very similar visual appearances. Distinguishing them is very challenge without textual cues.}
    \label{fig:Samples of drink}
\end{figure*}

To further verify the generality of the proposed method, we have collected a new image dataset named \textit{Drink Bottle dataset} which will be released soon. It consists of various types of drink bottle images contained in \textit{soft drink} and \textit{alcoholic drink} sets in ImageNet dataset \cite{russakovsky2015imagenet}. The dataset has 18,488 images divided into 20 categories: \textit{Root Beer, Ginger ale, Coca Cola, Pepsi Cola, Cream Soda, Egg Cream, Birch Beer, Tonic, Sarsaparilla, Orange Soda, Pulque, Kvass, Bitter, Guinness, Ouzo, Slivovitz, Drambuie, Vodka, Chablis and Sauterne}. The other settings follow Con-Text dataset. As shown in some image samples illustrated in Figure~\ref{fig:Samples of drink}, the images in this dataset have low-resolutions and dense words. Moreover, a lot of fonts are in exotic styles. All these aspects make the fine-grained classification on this dataset more challenge than on the Con-Text dataset.



\subsubsection{\textbf{Evaluation measures}}
The Average Precision (AP) of each category and the mean of AP (mAP) over all categories are used to measure the performance. Following the work in \cite{karaoglu2016words}, we first sort the classification score in descending order for each category, then the AP is given by:
\begin{equation}
AP = \sum_{k=1}^{n} P(k) * (R(k) - R(k-1)),
\end{equation}
where $P(k)$ and $R(k)$ stands for precision and respectively recall at cut-off $k$ in the sorted list, $R(0) = 0$, and $n$ is the number of images in current category.

\subsubsection{\textbf{Baselines}} \label{sssec:baselines}
We compare the proposed method with two baselines: a) Visual baseline. We fine tune the pre-trained GoogLeNet with a 28-category softmax classifier; b) Textual baseline. As described in Section~\ref{sssec:supervison over different features}, we have to add an extra class representing images without spotted text to train the network using only text information. Although we could evaluate the textual baseline using 29 categories, it is seriously imbalanced between the 28 original categories and the added category. Because most images in this dataset do not have text or the text reading system fails to spot words in images. Consequently, we train and test the textual baseline using 28 categories on a subset of images having spotted words. Since only text information is involved in the textual baseline, instead of using a weighted sum-pooling of word vectors described in Section~\ref{subsec:attention}, we take the classical average of the word vectors given by all spotted words as text feature for this baseline. For a fair comparison, we also train a visual baseline model and a model of the proposed method on this subset.

\subsubsection{\textbf{Implementation details}} \label{subsubsec:implementation}
Our method is built on several pre-trained networks: GoogLeNet~\footnote{\url{https://github.com/kmatzen/googlenet-caffe2torch}} \cite{szegedy2015going} to extract image features, TextBoxes~\footnote{\url{https://github.com/MhLiao/TextBoxes}} \cite{liao2016textboxes} to detect texts in images, incremental CNN~\footnote{\url{http://www.robots.ox.ac.uk/~vgg/research/text/}} \cite{jaderberg2014synthetic} to recognize the detected words, and GloVe~\footnote{\url{https://github.com/stanfordnlp/GloVe}} \cite{pennington2014glove} to generate the word embedding. During training, the mini-batch size for SGD is set to 64. The learning rate is set to 0.01 for the first 7K iterations, and then is divided by 10 every 10K iterations. The models are trained for up to 30K iterations. We use a weight decay of 0.0001 and a momentum of 0.9.

For the text representation described in Section \ref{sssec:text representation}, we only consider spotted words having more than 3 letters. Because the words having less than 3 letters are more likely to be false positives or represent irrelevant information for fine-grained classification.


The loss weights in Eq.~\eqref{fuse_loss} are set respectively to $\beta_1 = 1.0$, $\beta_2 = 0.5$, and $\beta_3 = 0.5$. In the training stage, we use classical dataset augmentation strategy \cite{szegedy2015going} with different scales and aspect ratios. More precisely, we re-size each training image to 255 $\times$ 255. Then, we randomly crop it with re-size scale in the interval $[0.08, 1]$ and aspect ratio in $[3/4, 4/3]$. Finally, the cropped patch is re-sized to 224 $\times$ 224. Then randomly horizontal flip is also used with a probability of 0.5.
In the testing stage, images are directly re-sized to 224 $\times$ 224 and the standard ten-crop testing~\cite{szegedy2015going} is adopted. 

The proposed method is implemented using Torch \cite{collobert2011torch7} deep learning framework. All the experiments are conducted on a NVIDIA GTX TitanX GPU. The materials to reproduce all the reported results will be released soon. 


\subsection{Importance of attention unit and multi-supervisions} \label{subsec:ablation}

\begin{table}[t]
\setlength\tabcolsep{1pt}
\renewcommand{\arraystretch}{1.2}
\centering  
\begin{tabular*}{6cm}{@{\extracolsep{\fill}}cc}
\hline
\textbf{Text fusion method} &\textbf{mAP(\%)} \\ \hline
Average-pooling &78.6 \\ 
Weighted sum-pooling &\textbf{79.6} \\  \hline
\end{tabular*}
\caption{Importance of semantic attention mechanism.} 
\label{table:effect of attention}
\end{table}

\begin{table}[t]
\setlength\tabcolsep{1pt}
\renewcommand{\arraystretch}{1.2}
\centering  
\begin{tabular*}{6cm}{@{\extracolsep{\fill}}cc}
\hline
\textbf{Supervision method} &\textbf{mAP(\%)} \\ \hline
Single-Supervision &78.9 \\ 
Multi-Supervisions &\textbf{79.6} \\  \hline
\end{tabular*}
\caption{Effect of multi-supervisions. In Single-Supervision, only a classifier trained on the combined feature $f_c$ is used. Multi-Supervisions denotes the use of the three classifiers described in Section~\ref{sssec:supervison over different features} respectively on $f_c$, $f_v$, and $f_a$ features.}
\label{table:effect of multi-supervisions}
\end{table}

\begin{table*}[h]
\setlength\tabcolsep{1pt}
\renewcommand{\arraystretch}{1.2}
\centering
\begin{tabular}{c | c c c c c c c c c c c c c c c c c c c c c c c c c c c c | c}
\hline
Method &\footnotesize{mas.} &\footnotesize{pet.} &\footnotesize{res.} &\footnotesize{com.} &\footnotesize{theat.} &\footnotesize{rep.} &\footnotesize{b.s.} &\footnotesize{bak.} &\footnotesize{med.} &\footnotesize{bar.} &\footnotesize{piz.} &\footnotesize{din.} &\footnotesize{h.p.} &\footnotesize{bist.} &\footnotesize{tea} &\footnotesize{sch.} &\footnotesize{pha.} &\footnotesize{fun.} &\footnotesize{cou.} &\footnotesize{tav.} &\footnotesize{motel} &\footnotesize{pack.} &\footnotesize{cafe} &\footnotesize{toba.} &\footnotesize{d.c.} &\footnotesize{disc.} &\footnotesize{steak} &\footnotesize{pawn} &\footnotesize{~mAP} \\ \hline
\begin{tabular}[c]{@{}c@{}}Visual\\ Baseline\textsuperscript{*}\end{tabular} &\scriptsize{60.4} &\scriptsize{57.8} &\scriptsize{30.4} &\scriptsize{63.4} &\scriptsize{88.2} &\scriptsize{46.6} &\scriptsize{76.9} &\scriptsize{59.9} &\scriptsize{61.3} &\scriptsize{66.7} &\scriptsize{38.6} &\scriptsize{74.8} &\scriptsize{90.4} &\scriptsize{17.9} &\scriptsize{30.2} &\scriptsize{71.0} &\scriptsize{30.8} &\scriptsize{51.4} &\scriptsize{60.1} &\scriptsize{43.4} &\scriptsize{80.3} &\scriptsize{53.1} &\scriptsize{23.8} &\scriptsize{43.4} &\scriptsize{50.6} &\scriptsize{11.1} &\scriptsize{21.7} &\scriptsize{42.7} &\scriptsize{51.7}  \\ \hline
\begin{tabular}[c]{@{}c@{}}Textual\\ Baseline\textsuperscript{*}\end{tabular} &\scriptsize{28.8} &\scriptsize{16.8} &\scriptsize{12.5} &\scriptsize{28.6} &\scriptsize{40.6} &\scriptsize{12.1} &\scriptsize{42.9} &\scriptsize{53.8} &\scriptsize{23.2} &\scriptsize{56.5} &\scriptsize{55.2} &\scriptsize{60.4} &\scriptsize{81.8} &\scriptsize{22.8} &\scriptsize{25.6} &\scriptsize{44.3} &\scriptsize{55.9} &\scriptsize{69.6} &\scriptsize{46.0} &\scriptsize{18.6} &\scriptsize{81.2} &\scriptsize{47.6} &\scriptsize{57.1} &\scriptsize{46.2} &\scriptsize{87.1} &\scriptsize{37.5} &\scriptsize{61.3} &\scriptsize{75.9} 
 &\scriptsize{46.1} \\ \hline
ours\textsuperscript{*} &\scriptsize{\textbf{74.3}} &\scriptsize{\textbf{72.7}} &\scriptsize{\textbf{46.7}} &\scriptsize{\textbf{79.7}} &\scriptsize{\textbf{93.5}} &\scriptsize{\textbf{60.0}} &\scriptsize{\textbf{90.6}} &\scriptsize{\textbf{84.0}} &\scriptsize{\textbf{81.3}} &\scriptsize{\textbf{92.6}} &\scriptsize{\textbf{84.3}} &\scriptsize{\textbf{88.3}} &\scriptsize{\textbf{90.9}} &\scriptsize{\textbf{32.6}} &\scriptsize{\textbf{54.7}} &\scriptsize{\textbf{85.6}} &\scriptsize{\textbf{78.9}} &\scriptsize{\textbf{90.9}} &\scriptsize{\textbf{83.3}} &\scriptsize{\textbf{63.1}} &\scriptsize{\textbf{98.1}} &\scriptsize{\textbf{85.0}} &\scriptsize{\textbf{72.0}} &\scriptsize{\textbf{77.7}} &\scriptsize{\textbf{96.1}} &\scriptsize{\textbf{51.9}} &\scriptsize{\textbf{82.8}} &\scriptsize{\textbf{91.9}} &\scriptsize{\textbf{78.0}}   \\ \hline
~\\[-8pt] \hline
\begin{tabular}[c]{@{}c@{}}Visual\\ Baseline\end{tabular} &\scriptsize{81.4} &\scriptsize{84.3} &\scriptsize{72.4} &\scriptsize{74.3} &\scriptsize{86.1} &\scriptsize{72.9} &\scriptsize{85.5} &\scriptsize{80.4} &\scriptsize{74.2} &\scriptsize{85.9} &\scriptsize{79.9} &\scriptsize{75.7} &\scriptsize{69.0} &\scriptsize{20.9} &\scriptsize{55.7} &\scriptsize{66.6} &\scriptsize{72.7} &\scriptsize{70.6} &\scriptsize{60.3} &\scriptsize{33.8} &\scriptsize{72.9} &\scriptsize{60.3} &\scriptsize{31.5} &\scriptsize{45.6} &\scriptsize{49.0} &\scriptsize{3.1} &\scriptsize{21.9} &\scriptsize{30.6} &\scriptsize{61.3} \\ \hline
ours &\scriptsize{\textbf{81.8}} &\scriptsize{\textbf{89.5}} &\scriptsize{\textbf{78.6}} &\scriptsize{\textbf{80.6}} &\scriptsize{\textbf{92.4}} &\scriptsize{\textbf{80.1}} &\scriptsize{\textbf{94.2}} &\scriptsize{\textbf{89.6}} &\scriptsize{\textbf{83.6}} &\scriptsize{\textbf{95.8}} &\scriptsize{\textbf{90.4}} &\scriptsize{\textbf{86.7}} &\scriptsize{\textbf{80.3}} &\scriptsize{\textbf{32.4}} &\scriptsize{\textbf{68.9}} &\scriptsize{\textbf{81.3}} &\scriptsize{\textbf{88.4}} &\scriptsize{\textbf{88.4}} &\scriptsize{\textbf{78.5}} &\scriptsize{\textbf{52.2}} &\scriptsize{\textbf{93.3}} &\scriptsize{\textbf{85.0}} &\scriptsize{\textbf{57.0}} &\scriptsize{\textbf{72.3}} &\scriptsize{\textbf{93.3}} &\scriptsize{\textbf{51.7}} &\scriptsize{\textbf{74.3}} &\scriptsize{\textbf{87.0}} &\scriptsize{\textbf{79.6}}  \\  \hline
~\\[-6pt] \hline
\begin{tabular}[c]{@{}c@{}}\scriptsize{Karaoglu et al.}\\ \cite{karaoglu2013text}\end{tabular} &\scriptsize{34.9} &\scriptsize{45.2} &\scriptsize{51.1} &\scriptsize{33.8} &\scriptsize{48.5} &\scriptsize{17.8} &\scriptsize{60.0} &\scriptsize{37.8} &\scriptsize{48.5} &\scriptsize{55.2} &\scriptsize{55.2} &\scriptsize{43.4} &\scriptsize{65.7} &\scriptsize{9.1} &\scriptsize{12.5} &\scriptsize{44.3} &\scriptsize{60.8} &\scriptsize{43.0} &\scriptsize{35.2} &\scriptsize{10.6} &\scriptsize{53.0} &\scriptsize{38.2} &\scriptsize{16.2} &\scriptsize{29.0} &\scriptsize{50.9} &\scriptsize{18.7} &\scriptsize{28.1} &\scriptsize{44.5} &\scriptsize{39.0}\\  \hline
\begin{tabular}[c]{@{}c@{}}\scriptsize{Karaoglu et al.}\\ \cite{karaoglu2016words}\end{tabular} &\scriptsize{-} &\scriptsize{-} &\scriptsize{-} &\scriptsize{-} &\scriptsize{-} &\scriptsize{-} &\scriptsize{-} &\scriptsize{-} &\scriptsize{-} &\scriptsize{-} &\scriptsize{-} &\scriptsize{-} &\scriptsize{-} &\scriptsize{-} &\scriptsize{-} &\scriptsize{-} &\scriptsize{-} &\scriptsize{-} &\scriptsize{-} &\scriptsize{-} &\scriptsize{-} &\scriptsize{-} &\scriptsize{-} &\scriptsize{-} &\scriptsize{-} &\scriptsize{-} &\scriptsize{-} &\scriptsize{-} &\scriptsize{77.3}\\ \hline
\end{tabular}
\caption{Classification performance for baselines, our method, and two state-of-the art methods on the Con-Text dataset. Models with $*$ are trained and tested on the subset of images containing spotted texts. For clearness, the words in the first row are \textit{massage parlor, pet shop, restaurant, computer store, theatre, repair shop, bookstore, bakery, medical center, barbershop, pizzeria, diner, hotspot, bistro, teahouse, school, pharmacy, funeral, country store, tavern, motel, packinghouse, cafe, tobacco shop, dry cleaner, discount house, steakhouse}, and \textit{pawnshop} respectively.}
\label{table:performance over baselines}
\end{table*}

\begin{table*}[h]
\setlength\tabcolsep{1pt}
\renewcommand{\arraystretch}{1.2}
\centering  
\begin{tabular}{c | c c c c c c c c c c c c c c c c c c c c | c}
\hline
Method &\small{~r.b.} &\small{~g.a.} &\small{~coke} &\small{~pepsi} &\small{~c.s.} &\small{~e.c.} &\small{~b.c.} &\small{~tonic} &\small{~sarsap.} &\small{~o.s.} &\small{~pulque} &\small{~kvass} &\small{~bitter} &\small{~guin.} &\small{~ouzo} &\small{~sliv.} &\small{~dramb.} &\small{~vodka} &\small{~chabl.} &\small{~saut.~} &\small{~mAP} \\ \hline
\begin{tabular}[c]{@{}c@{}}Visual\\ Baseline\textsuperscript{*}\end{tabular} &\scriptsize{64.1} &\scriptsize{63.2} &\scriptsize{94.8} &\scriptsize{\textbf{98.1}} &\scriptsize{27.5} &\scriptsize{40.4} &\scriptsize{33.1} &\scriptsize{59.6} &\scriptsize{29.5} &\scriptsize{77.8} &\scriptsize{58.3} &\scriptsize{26.4} &\scriptsize{76.4} &\scriptsize{97.0} &\scriptsize{64.8} &\scriptsize{47.3} &\scriptsize{88.4} &\scriptsize{82.2} &\scriptsize{90.9} &\scriptsize{87.1} &\scriptsize{65.4}   \\ \hline
\begin{tabular}[c]{@{}c@{}}Textual\\ Baseline\textsuperscript{*}\end{tabular} &\scriptsize{54.7} &\scriptsize{61.9} &\scriptsize{34.1} &\scriptsize{76.5} &\scriptsize{27.0} &\scriptsize{0.2} &\scriptsize{22.3} &\scriptsize{33.6} &\scriptsize{18.6} &\scriptsize{54.5} &\scriptsize{37.4} &\scriptsize{23.2} &\scriptsize{40.2} &\scriptsize{90.7} &\scriptsize{58.7} &\scriptsize{9.5} &\scriptsize{75.6} &\scriptsize{61.5} &\scriptsize{64.2} &\scriptsize{55.7} &\scriptsize{45.0}  \\ \hline
ours\textsuperscript{*} &\scriptsize{\textbf{80.8}} &\scriptsize{\textbf{74.0}} &\scriptsize{\textbf{96.1}} &\scriptsize{98.0} &\scriptsize{\textbf{54.4}} &\scriptsize{\textbf{41.1}} &\scriptsize{\textbf{47.4}} &\scriptsize{\textbf{69.1}} &\scriptsize{\textbf{48.9}} &\scriptsize{\textbf{85.4}} &\scriptsize{\textbf{65.6}} &\scriptsize{\textbf{40.2}} &\scriptsize{\textbf{84.7}} &\scriptsize{\textbf{98.5}} &\scriptsize{\textbf{79.8}} &\scriptsize{\textbf{59.5}} &\scriptsize{\textbf{91.7}} &\scriptsize{\textbf{90.8}} &\scriptsize{\textbf{93.5}} &\scriptsize{\textbf{90.9}} &\scriptsize{\textbf{74.5}}   \\ \hline
~\\[-8pt] \hline
\begin{tabular}[c]{@{}c@{}}Visual\\ Baseline\end{tabular} &\scriptsize{61.9} &\scriptsize{54.2} &\scriptsize{94.1} &\scriptsize{95.6} &\scriptsize{32.4} &\scriptsize{60.7} &\scriptsize{35.7} &\scriptsize{53.4} &\scriptsize{41.1} &\scriptsize{78.5} &\scriptsize{51.7} &\scriptsize{31.6} &\scriptsize{66.5} &\scriptsize{93.3} &\scriptsize{45.3} &\scriptsize{43.6} &\scriptsize{75.3} &\scriptsize{75.9} &\scriptsize{85.1} &\scriptsize{84.4} &\scriptsize{63.1}  \\ \hline
ours &\scriptsize{\textbf{76.8}} &\scriptsize{\textbf{70.3}} &\scriptsize{\textbf{95.8}} &\scriptsize{\textbf{96.9}} &\scriptsize{\textbf{52.5}} &\scriptsize{\textbf{73.2}} &\scriptsize{\textbf{42.9}} &\scriptsize{\textbf{65.6}} &\scriptsize{\textbf{57.8}} &\scriptsize{\textbf{87.4}} &\scriptsize{\textbf{61.7}} &\scriptsize{\textbf{39.4}} &\scriptsize{\textbf{77.3}} &\scriptsize{\textbf{96.7}} &\scriptsize{\textbf{63.8}} &\scriptsize{\textbf{54.1}} &\scriptsize{\textbf{79.9}} &\scriptsize{\textbf{85.7}} &\scriptsize{\textbf{90.2}} &\scriptsize{\textbf{88.7}} &\scriptsize{\textbf{72.8}}  \\  \hline
\end{tabular}
\caption{Classification performance for baselines and the proposed method on the Drink Bottle dataset. The category of every logogram in this table is depicted in Section~\ref{sssec:datasets}. Similar to Table~\ref{table:performance over baselines}, models with $*$ are trained and tested on the subset of images containing spotted texts.}
\label{table:performance on drink bottles}
\end{table*}

\subsubsection{\textbf{Importance of attention unit}} \label{sssec:the influence of attention unit}


The state-of-the-art text reading system may result in some noisy texts including words that are irrelevant to the scene and incorrectly spotted words. 
These words may be similar in appearance but semantically far away, such as `hotpot' and `hotspot', `barbera' and `barber', and etc. We propose to use the attention mechanism to assign appropriate weights to different spotted words. The weighted sum-pooling of word vectors is then used to represent the textual information related to the scene. As depicted in Table~\ref{table:effect of attention} on the Con-Text dataset, the proposed attention-based method achieves nearly 1\% performance gain over the classical average-pooling scheme.

Some weights given by the attention mechanism over spotted words in several images are given in Figure~\ref{fig:visualization of results}. The proposed attention unit can accurately select the relevant words and eliminate irrelevant ones, even for the failure cases in the third row. The attended weights are reasonable. Their visual features being very close to the incorrectly predicted classes are responsible for their fails. Some failure cases (fourth row) are due to the performance of the text reading system. The relevant words are not detected or incorrectly recognized, such as `LAUNDROMAT' in (m), `TEA' in (n), `ROOT BEER' on the right can in (o), and `SODA' in (p). Yet the words having large weights are semantically close to the scene class.  


\subsubsection{\textbf{Effect of multi-supervisions}}

Although the main idea in this paper is to use combined text and visual feature $f_c$ to improve fine-grained classification, we rely on multi-supervisions on $f_c$, $f_v$, and $f_a$ respectively during the training stage to boost the performance of the proposed classifier based on $f_c$.
Without adjusting carefully the weights over different supervisions in Eq.~\eqref{fuse_loss}, we simply set the loss weights for combined text and visual classifier, visual classifier, and text classifier as described in Section~\ref{subsubsec:implementation}. As depicted in Table~\ref{table:effect of multi-supervisions} on the Con-Text dataset, the multi-supervisions provide about a 0.7\% performance gain compared to single-supervision.
Note that the two extra supervision classifiers are only used in training stage. During test stage, we simply use the classifier based on combined visual and text feature $f_c$. Consequently, the 0.7\% performance gain thanks to multi-supervisions is achieved without any loss in runtime during the test.

\subsection{Comparison with baselines} \label{subsec:compbaseline}

We compare the proposed method with several baseline methods described in Section~\ref{sssec:baselines}. We have conducted two types of experiments using two different dataset settings: 1) Visual baseline and the proposed method trained on all training images and tested on all testing images; 2) Visual baseline, textual baseline, and the proposed method trained on the subset of training images and tested on the subset of testing images having spotted texts in each dataset. We add $*$ after corresponding name to distinguish these experiments from previous ones.

Table~\ref{table:performance over baselines} displays the quantitative comparisons on the Con-Text dataset.
In general, the textual baseline achieves results on par with visual baseline on images having spotted texts in this dataset. This shows that the text feature is as useful as visual feature for fine-grained classification on images containing texts. The proposed method using the combination of text and visual features significantly outperforms the baseline based only on text or visual feature. A large improvement is observed particularly for the images having similar visual layout but discriminative textual cues, e.g., \textit{Dry Cleaner}, \textit{Discount House},\textit{Steak House}, and \textit{Pawn Shop}. For the images having fewer texts or no texts, the improvement is less significant. The observations on the Con-Text dataset also hold for the Drink Bottle dataset. As depicted in Table~\ref{table:performance on drink bottles}, the proposed method also significantly outperforms the visual and textual baselines. Yet, the improvement over visual baseline (about 10\%) is less important as compared to the improvement (about 20\%) on the Con-Text dataset. The main reason is likely to be that recognition of the exotic texts in the low-resolution drink bottle images is more challenge. The text reading system would yield more noisy texts. The fact that less accurate text recognition gives less significant improvements also demonstrates the importance and interestingness of integrating text appearances into visual appearances for fine-grained classification.



To further analyze the benefit of integrating text information into visual appearance for fine-grained classification, we compare the performance of the proposed method and visual baseline on the subset of testing images having spotted texts and subset without texts separately in Con-Text dataset. In this dataset, about 42\% images contain texts. The quantitative comparison is given in Table~\ref{table:detail results}. The proposed method improves dramatically the performance on images having texts as observed in the above experiments. Surprisingly, It also boosts slightly the performance for images without texts. One reason for this improvement may lie on the multi-supervisions of the network using shared parameters. 


\begin{table}[htbp]
\setlength\tabcolsep{1pt}
\renewcommand{\arraystretch}{1.2}
\centering  
\begin{tabular*}{7cm}{@{\extracolsep{\fill}}ccc}
\hline
  &image with texts &image without texts \\ \hline
Visual Baseline &52.8 &65.8 \\ 
ours &\textbf{79.0} &\textbf{71.7} \\  \hline
\end{tabular*}
\caption{Average precision (in \%) on the subset of images having spotted words and the rest images in the Con-Text dataset. Both models are trained on all training images.}
\label{table:detail results}
\end{table}

\subsection{Comparison with other methods} \label{subsec:comparison}

We compare the proposed method with the state-of-the-art methods~\cite{karaoglu2013text,karaoglu2016words} on the Con-Text dataset. In~\cite{karaoglu2013text,karaoglu2016words}. As shown in Table~\ref{table:performance over baselines}, the proposed method improves the state-of-the-art results from 77.3\% to 79.6\% mAP. Besides, the method in~\cite{karaoglu2016words} relies on both global visual information and local object information given by 100 EdgeBoxes~\cite{zitnick2014edge}. The proposed method only relies on global visual information without the additional local part-based visual features. Moreover, the proposed method is end-to-end trainable, which is more elegant and efficient. 


\begin{figure*}[htbp]
    \centering
    \subfloat[\scriptsize{BARBERSHOP} \textcolor{blue}{BARBERSHOP}\newline BARBER: 1\newline SHOP: 7.8e-7\newline MENUS: 2.8e-8\newline ROOM: 1.2e-11 \newline BARBS: 3.8e-18]{\includegraphics[width=1.6in, height=1.25in]{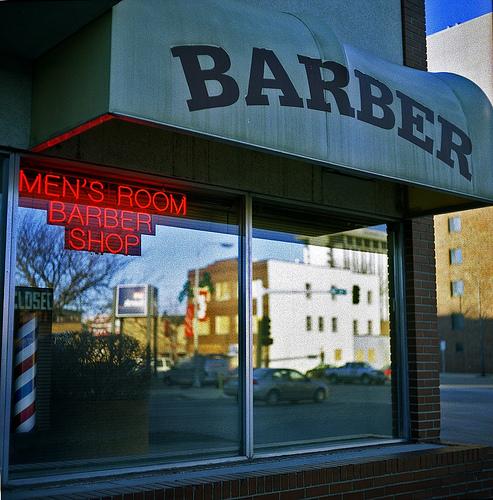}%
        \label{fig_first_case}}
    \hfil
    \subfloat[\scriptsize{CAFE} \textcolor{blue}{CAFE}\newline COFFEE: 0.97\newline ESPRESSO: 0.03 \newline CAPPUCCINO: 2.0e-10 \newline ITALIAN: 2.2e-12]{\includegraphics[width=1.6in, height=1.25in]{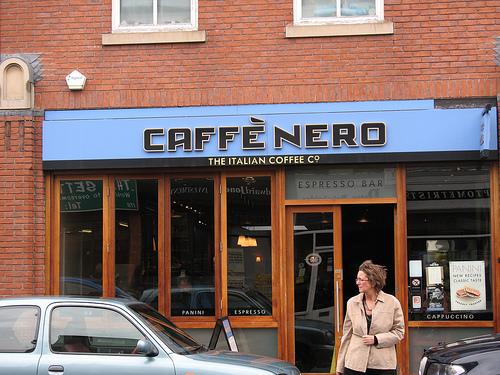}%
        \label{fig_second_case}}
    \hfil
    \subfloat[\scriptsize{BAKERY} \textcolor{blue}{BAKERY}\newline CAKES: 0.57\newline PASTRIES: 0.43\newline OPEN: 5.5e-9\newline EGGO: 1.1e-10\newline DANISH: 3.1e-11]{\includegraphics[width=1.6in, height=1.25in]{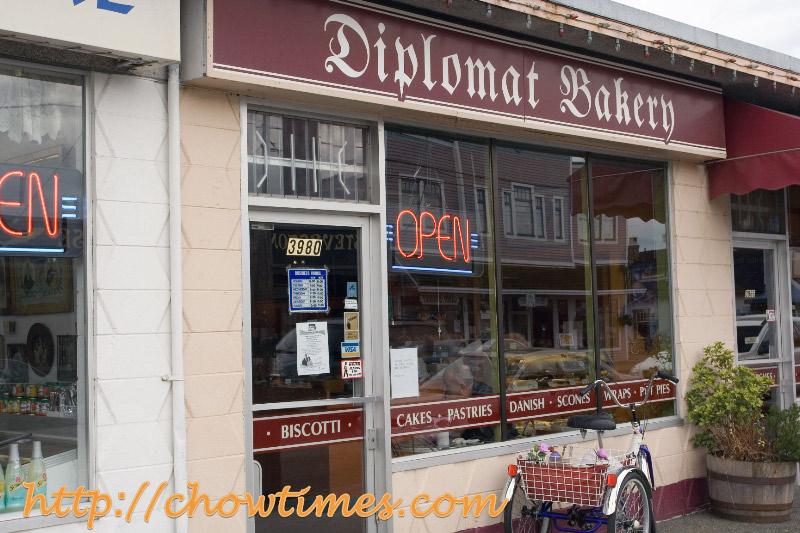}%
        \label{fig_third_case}}
    \hfil
    \subfloat[\scriptsize{CAFE} \textcolor{blue}{CAFE}\newline STARBUCKS: 1\newline SCOFF: 1.1e-8]{\includegraphics[width=1.6in, height=1.25in]{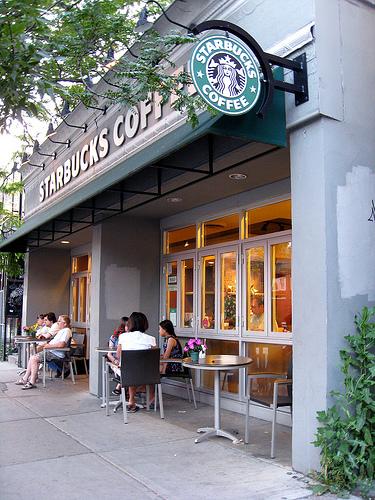}%
        \label{fig_fourth_case}}
    \hfil

    
    \subfloat[\scriptsize{ROOTBEER} \textcolor{blue}{ROOTBEER}\newline ROOT: 0.89\newline BEER: 0.11\newline BREWED: 1.3e-6\newline PURE: 2.4e-7\newline MICRO: 1.1e-9\newline MADE: 3.8e-10\newline NATURAL: 2.7e-11\newline RICH: 1.8e-11\newline EFL: 5.5e-12]{\includegraphics[width=1.6in, height=1.25in]{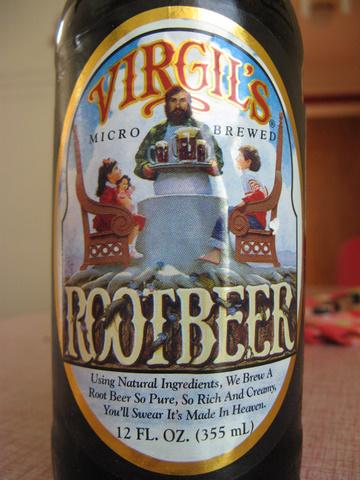}%
    	\label{fig_five_case}}
    \hfil
    \subfloat[\scriptsize{CHABLIS} \textcolor{blue}{CHABLIS}\newline CHABLIS: 0.99\newline FRANCE: 8.7e-12\newline FRANC: 1.1e-12\newline YIN: 2.4e-16\newline CON: 2.3e-18\newline CONTROL: 1.9e-18\newline BOUTIQUE: 2.5e-19\newline AFFILIATION: 6.2e-20]{\includegraphics[width=1.6in, height=1.25in]{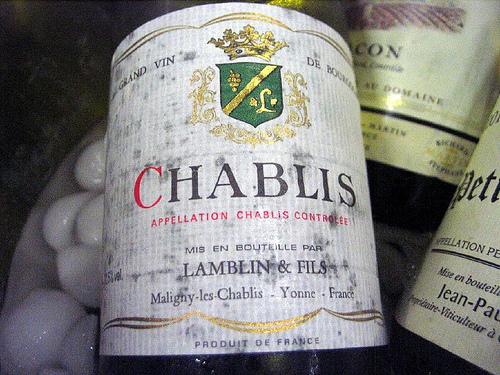}%
    	\label{fig_six_case}}
    \hfil
    \subfloat[\scriptsize{BITTER} \textcolor{blue}{BITTER}\newline BITTER: 0.99\newline BROWN: 4.05e-5\newline PREMIUM: 3.5e-9\newline SPECIAL: 2.8e-9\newline ENGLISH: 9.4e-11\newline EXTRA: 6.11e-11]{\includegraphics[width=1.6in, height=1.25in]{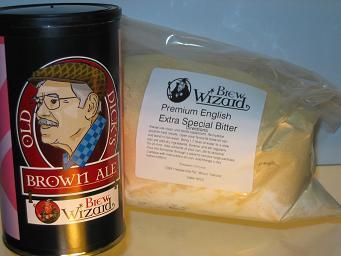}%
    	\label{fig_seven_case}}
    \hfil
    \subfloat[\scriptsize{GUINNESS} \textcolor{blue}{GUINNESS}\newline GUINNESS: 1\newline SPECIAL: 1.6e-25\newline EXPORT: 6.4e-27\newline QUINES: 1.3e-30]{\includegraphics[width=1.6in, height=1.25in]{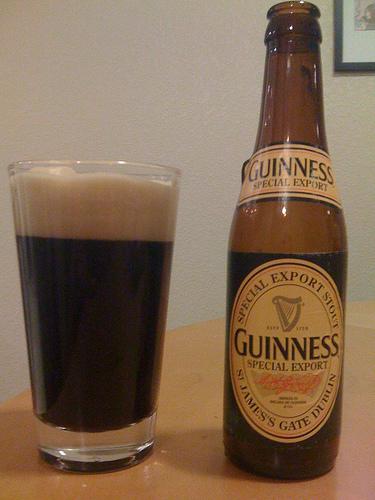}%
    	\label{fig_eight_case}}
    \hfil
    
    \subfloat[\scriptsize{PETSHOP} \textcolor{red}{COUNTRYSTORE}\newline PET: 0.56\newline STORE: 0.44\newline THE: 7.4e-13\newline VILLAGE: 5.5e-14]{\includegraphics[width=1.6in, height=1.25in]{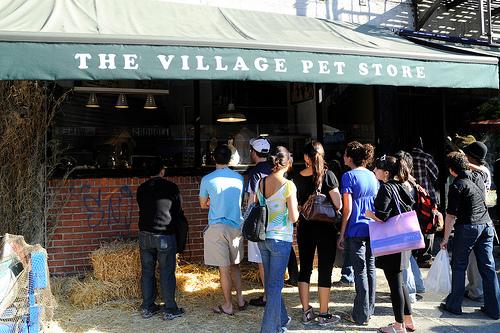}%
    	\label{fig_twelve_case}}
    \hfil
    \subfloat[\scriptsize{FUNERAL} \textcolor{red}{MOTEL}\newline CHAPEL: 0.50\newline MEMORIAL: 0.50\newline WOODEN: 1.5e-09]{\includegraphics[width=1.6in, height=1.25in]{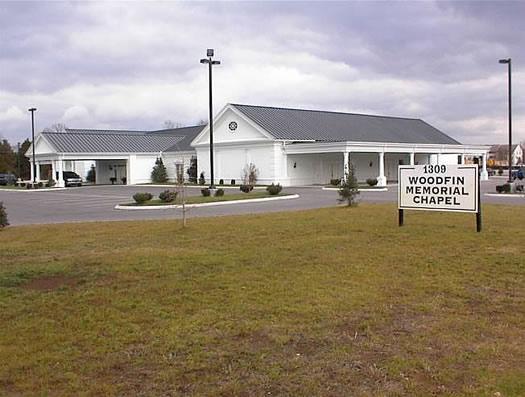}%
    	\label{fig_nine_case}}
    \hfil
    \subfloat[\scriptsize{CREAMSODA} \textcolor{red}{BIRCHBEER}\newline \scriptsize{SWEETENED}: 0.93\newline CREME: 0.07\newline OUNCES: 2.6e-5]{\includegraphics[width=1.6in, height=1.25in]{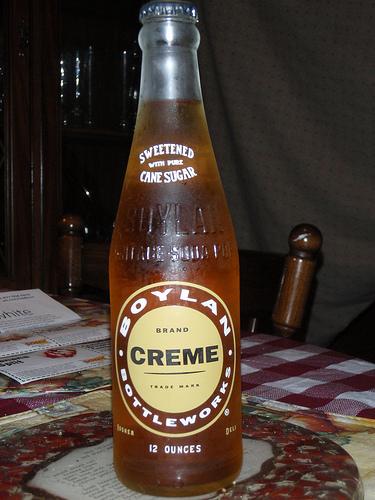}%
    	\label{fig_fifteen_case}}
    \hfil
    \subfloat[\scriptsize{GUINNESS} \textcolor{red}{BITTER}\newline GUINNESS: 0.50\newline BITTER: 0.50]{\includegraphics[width=1.6in, height=1.25in]{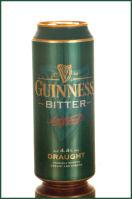}%
    	\label{fig_sixteen_case}} 
    \hfil
    
    \subfloat[\scriptsize{DRYCLEANER} \textcolor{red}{TOBACCOSHOP}\newline CIGARETTES: 0.91\newline DRINKS: 0.07\newline COLD: 0.02\newline DIAL: 9.7e-05\newline ACACIA: 1.5e-05\newline PLAY: 4.3e-08\newline WIN: 5.6e-09\newline VIDEO: 2.8e-11]{\includegraphics[width=1.6in, height=1.25in]{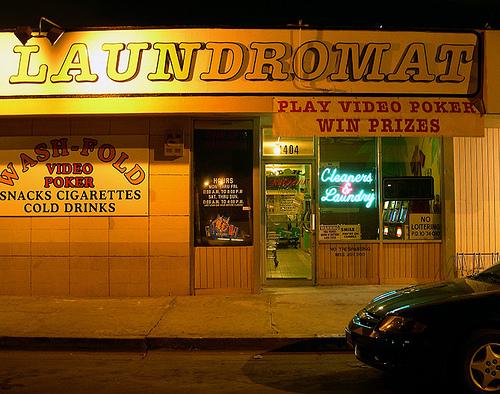}%
    	\label{fig_ten_case}}
    \hfil
    \subfloat[\scriptsize{TEAHOUSE} \textcolor{red}{CAFE}\newline CAFE: 1\newline ROOMS: 2.9e-06\newline LEA: 5.4e-09]{\includegraphics[width=1.6in, height=1.25in]{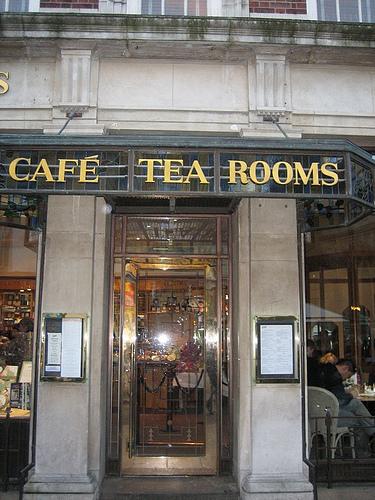}%
    	\label{fig_eleven_case}}
    \hfil
    \subfloat[\scriptsize{ROOTBEER} \textcolor{red}{GINGERALE}\newline GINGER: 0.76\newline ALE: 0.24\newline CAFFEINE: 1.03e-14]{\includegraphics[width=1.6in, height=1.25in]{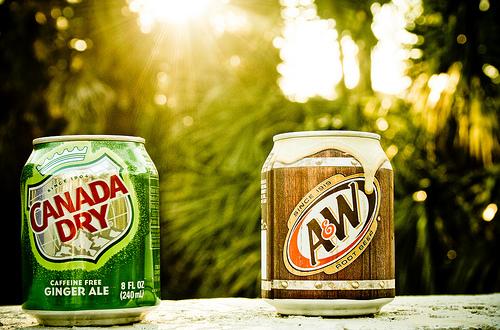}%
    	\label{fig_thirteen_case}}
    \hfil
    \subfloat[\scriptsize{CREAMSODA} \textcolor{red}{ROOTBEER}\newline ROOTBEER: 0.97\newline CAFFEINE: 0.02\newline COM: 0.01\newline FACTS: 3.0e-4\newline NUTRITION: 3.9e-8]{\includegraphics[width=1.6in, height=1.25in]{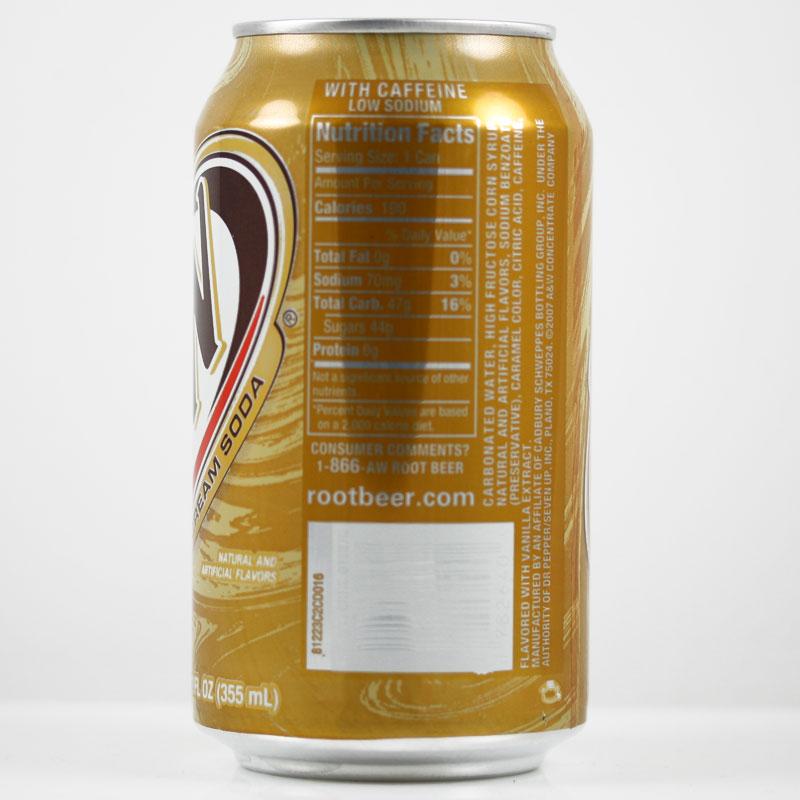}%
    	\label{fig_fourteen_case}}
    \hfil
    \caption{Some classification results on the Con-Text dataset and Drink Bottle dataset (better visualization in electronic verison). The ground truth and predicted class (in blue and red for correct and respectively incorrect predictions) are given in the first line at the bottom of each image. The spotted words and their corresponding weights are shown in the rest lines.}
    \label{fig:visualization of results}
\end{figure*}

Some fine-grained classification results of the proposed method are given in~Figure~\ref{fig:visualization of results}. The proposed attention mechanism accurately selects the most relevant words related the scene. From (a) to (h), the proposed method makes correct predictions regardless the very close visual appearances of different fine-grained classes. Some failure cases are also shown in the third and fourth row. Though relevant spotted words are highlighted by the proposed attention mechanism in the third row, the images are incorrectly classified due to their visual features. In the last row, the most related words are not spotted by the text reading system. The proposed method fails even if the semantically close words are highlighted by the attention mechanism. A better performance could be expected with a more robust text reading system in the future.


\subsection{Application to Product Retrieval} \label{subsec:product}
we present an application of the proposed framework for \textit{Product Retrieval} to further demonstrate its potential. Compared to image classification, image retrieval attracts more attention in e-commerce. Given a query image $I$, the objective of product retrieval is to select all product images from an image set withing the same category as $I$. We have conducted this experiment on the Drink Bottle dataset. We simply extract the input vector of the softmax-based classifier in fine-grained classification model. This input vector is considered as product feature for retrieval. The cosine similarity is used to rank images during matching. Following FlickrLogos-32's evaluation settings~\cite{romberg2011scalable}, the test set and train set are used as queries and respectively search set. We evaluate this application based on mAP. As shown in Table~\ref{table:performance of retrieval on drink bottles}, the proposed method boosts consistently and significantly the retrieval performance. 



\begin{table}[t]
\setlength\tabcolsep{2pt}
\renewcommand{\arraystretch}{1.2}
\centering  
\begin{tabular*}{8cm}{@{\extracolsep{\fill}}cc}
\hline
Method &mAP(\%) \\ \hline
Visual Baseline\textsuperscript{*} &60.0\\ 
Textual Baseline\textsuperscript{*} &28.9\\
ours\textsuperscript{*} &\textbf{72.6}\\ \hline
~\\[-8pt] \hline
Visual Baseline &48.0 \\ 
ours &\textbf{60.8} \\  \hline
\end{tabular*}
\caption{Evaluation of the proposed method for product retrieval on Drink Bottles dataset. Models with $*$ are trained and tested on the subset of images containing texts (see also Section~\ref{subsec:compbaseline}).}
\label{table:performance of retrieval on drink bottles}
\end{table}

\section{Conclusion} \label{sec: conclusion}

In this paper, we have introduced a unified, end-to-end trainable framework which integrates visual and text cues for fine-grained image classification. We have also incorporated the attention mechanism to better exploit the extracted text in images by only selecting the most relevant words. Using several state of the art components in an effective pipeline,  our experimental results on two datasets have demonstrated both the value and performance of the proposed method. Furthermore, the product retrieval experiment shows that the learned representation provides a potentially powerful tool for product search. It is worth noting that while it certainly makes sense to employ the state of the art in scene text and visual recognition as the components, our main motivation and contribution are to fill a gap in fine-grained image classification by exploiting multi-modal complementary information when present, and to do so with an end-to-end trainable neural network with the attention mechanism for implicitly exploiting the semantic relationships between textual and visual cues. The main drawback of using textual cues in the proposed framework for fine-grained image classification is clearly the absence of texts in some images. Besides the scene texts in images, the additional textual cues could be user tags, images attributes, and so on. In the future, we would like to incorporate other textual cues to handle the problem of no texts in images.

\ifCLASSOPTIONcaptionsoff
  \newpage
\fi






\bibliographystyle{IEEEtran}

\bibliography{reference}

%








\end{document}